\documentclass[10pt,letterpaper]{article}
\usepackage[top=0.85in,left=2.75in,footskip=0.75in]{geometry}

\usepackage{amsmath,amssymb}

\usepackage{changepage}

\usepackage[utf8x]{inputenc}

\usepackage{textcomp,marvosym}

\usepackage{cite}

\usepackage{nameref,hyperref}

\usepackage[right]{lineno}

\usepackage{microtype}
\DisableLigatures[f]{encoding = *, family = * }

\usepackage[table]{xcolor}

\usepackage{array}

\usepackage{graphicx}
\usepackage{amsmath}
\usepackage{amssymb}
\usepackage{color}
\usepackage{algorithm}
\usepackage{algpseudocode}

\newcommand{\ssse}{\sqsubseteq}

\newcommand{\prm}{\prime}

\newcommand{\bg}[0]{\boldsymbol{g}}

\newcommand{\bq}[0]{\boldsymbol{q}}

\newcommand{\bw}[0]{\boldsymbol{w}}
\newcommand{\bx}[0]{\boldsymbol{x}}

\newcommand{\argmin}{\mathop{\rm argmin}}
\newcommand{\eq}[1]{(\ref{#1})}

\newcommand{\RR}{\mathbb{R}}

\newcommand{\cA}{{\cal A}}

\newcommand{\cG}{{\cal G}}

\newcommand{\cQ}{{\cal Q}}

\newcommand{\cS}{{\cal S}}

\usepackage{bm}
\usepackage{placeins}

\usepackage{hyperref}
\usepackage{booktabs}
\usepackage{colortbl}
\usepackage[normalem]{ulem}
\usepackage[toc,page]{appendix}

\newcolumntype{+}{!{\vrule width 2pt}}

\newlength\savedwidth

\newcommand\thickhline{\noalign{\global\savedwidth\arrayrulewidth\global\arrayrulewidth 2pt}%
\hline
\noalign{\global\arrayrulewidth\savedwidth}}

\usepackage{setspace} 
\raggedright
\usepackage[aboveskip=1pt,labelfont=bf,labelsep=period,justification=raggedright,singlelinecheck=off]{caption}

\makeatletter
\renewcommand{\@biblabel}[1]{\quad#1.}
\makeatother

\usepackage{lastpage,fancyhdr,graphicx}
\usepackage{epstopdf}
\fancyheadoffset[L]{2.25in}
\fancyfootoffset[L]{2.25in}
\begin{document}
\vspace*{0.2in}

\begin{flushleft}
{\Large
\textbf\newline{Supervised sequential pattern mining of event sequences in sport to identify important patterns of play: an application to rugby union} % Please use "sentence case" for title and headings (capitalize only the first word in a title (or heading), the first word in a subtitle (or subheading), and any proper nouns).
}
\newline
\\
Rory Bunker\textsuperscript{1*},
Keisuke Fujii\textsuperscript{1,2},
Hiroyuki Hanada\textsuperscript{1},
Ichiro Takeuchi\textsuperscript{1,3}
\\
\bigskip
\textbf{1} RIKEN Center for Advanced Intelligence Project, Tokyo, Japan
\\
\textbf{2} Graduate School of Informatics, Nagoya University, Nagoya, Aichi, Japan
\\
\textbf{3} Department of Computer Science, Nagoya Institute of Technology, Nagoya, Aichi, Japan
\\
\bigskip

% Insert additional author notes using the symbols described below. Insert symbol callouts after author names as necessary.
% 
% Remove or comment out the author notes below if they aren't used.
%
% Primary Equal Contribution Note
%\Yinyang These authors contributed equally to this work.

% Additional Equal Contribution Note
% Also use this double-dagger symbol for special authorship notes, such as senior authorship.
%\ddag These authors also contributed equally to this work.

%\textcurrency Current Address: Dept/Program/Center, Institution Name, City, State, Country % change symbol to "\textcurrency a" if more than one current address note
% \textcurrency b Insert second current address 
% \textcurrency c Insert third current address


% Group/Consortium Author Note
%\textpilcrow Membership list can be found in the Acknowledgments section.

% Use the asterisk to denote corresponding authorship and provide email address in note below.
* Corresponding author \newline Email: rorybunker@gmail.com

\end{flushleft}
\section*{Abstract}
%%%
Given a set of sequences comprised of time-ordered events, sequential pattern mining is useful to identify frequent subsequences from different sequences or within the same sequence. 
However, in sport, these techniques cannot determine the importance of particular patterns of play to good or bad outcomes, which is often of greater interest to coaches and performance analysts.
In this study, we apply a recently proposed supervised sequential pattern mining algorithm called safe pattern pruning (SPP) to 490 labelled event sequences representing passages of play from one rugby team’s matches from the 2018 Japan Top League.
We compare the SPP-obtained patterns that are the most discriminative between scoring and non-scoring outcomes from both the team's and opposition teams' perspectives, with the most frequent patterns obtained with well-known unsupervised sequential pattern mining algorithms when applied to subsets of the original dataset, split on the label.
Our obtained results found that linebreaks, successful lineouts, regained  kicks  in  play, repeated phase-breakdown play, and failed exit plays by the opposition team were identified as as the patterns that discriminated most between the team scoring and not scoring.
Opposition team linebreaks, errors made by the team, opposition team lineouts, and repeated phase-breakdown play by the opposition team were identified as the patterns that discriminated most between the opposition team scoring and not scoring.
It was also found that, by virtue of its supervised nature as well as its pruning and safe-screening properties, SPP obtained a greater variety of generally more sophisticated patterns than the unsupervised models that are likely to be of more utility to coaches and performance analysts.
%%%

% Please keep the Author Summary between 150 and 200 words
% Use first person. PLOS ONE authors please skip this step. 
% Author Summary not valid for PLOS ONE submissions.   
%\section*{Author summary}
%Lorem ipsum dolor sit amet, consectetur adipiscing elit. Curabitur eget porta erat. Morbi consectetur est vel gravida pretium. Suspendisse ut dui eu ante cursus gravida non sed sem. Nullam sapien tellus, commodo id velit id, eleifend volutpat quam. Phasellus mauris velit, dapibus finibus elementum vel, pulvinar non tellus. Nunc pellentesque pretium diam, quis maximus dolor faucibus id. Nunc convallis sodales ante, ut ullamcorper est egestas vitae. Nam sit amet enim ultrices, ultrices elit pulvinar, volutpat risus.

%\linenumbers
\section*{Introduction}

%%%
Large amounts of data are now being captured in sport as a result of the increased use of GPS tracking and video analysis systems, as well as enhancements in computing power and storage, and there is great interest in making use of this data for performance analysis purposes.
A wide variety of methods have been used in the analysis of sports data, ranging from statistical methods to, more recently, machine learning and data mining techniques.
%%%

%%%
Among the various analytical frameworks available in sports analytics, in this paper, we adopt an approach to extract events from sports matches and analyze sequences of events.
The most basic events-based approach is based on the analysis of the \emph{frequencies} of events.
These frequencies can be used as performance indicators \cite{hughes2002use} by comparing the frequency of each event in positive outcomes (winning, scoring points, etc.) and negative outcomes (losing, conceding points, etc.) in order to investigate which events are commonly associated with these outcomes.
However, frequency-based analyses have drawbacks in that the information contained in the order of events cannot be exploited.
%%%

%%%
In this study, we consider a sequence of events, and refer to a partial sequence of events a \emph{sequential pattern} or simply a \emph{pattern (of play)}.
In sports, the occurrence of certain events in a particular order often has a strong influence on outcomes, so it is useful to use patterns as a basic analytical unit.
Invasion sports such as rugby (as well as soccer and basketball, for example) have many events and patterns that occur very frequently while having a paucity of events that are important for scoring.
For instance, in soccer, a pattern consisting of an accurate cross followed by a header that is on target will occur much less frequently than a pattern consisting of repeated passes between players, but the former pattern is likely to be of much greater interest to coaches and performance analysts because there is a good chance that the pattern may lead to a goal being scored. 
%%%

%%%
The computational framework for finding patterns from sequential data that have specific characteristics is known as \emph{sequential mining} in the field of data mining.
The most basic problem setup in sequential mining is to enumerate frequent patterns, which is called \emph{frequent sequential mining}.
Although the total number of patterns (i.e., the number of ordered sequences of all possible events) is generally very large, it is possible to efficiently enumerate patterns that appear more than a certain frequency by making effective use of branch-and-bound techniques.
Frequent sequential mining is categorised as an \emph{unsupervised learning technique} in the terminology of machine learning.
%%%

%%%
When applying frequent sequential mining to data from sport, there are several options.
The first option is to simply extract the frequent patterns from the entire dataset.
The drawback of this approach is that it is not possible to distinguish whether a pattern leads to good or bad outcomes.
The second option is to split the dataset into a ``good-outcome" dataset and a ``bad-outcome" dataset, and perform frequent sequential mining on each dataset.
The third option is to perform frequent sequential mining on the entire dataset to identify frequent patterns, and then create a machine learning model that uses the patterns as features to predict whether the outcomes are good or bad.
The disadvantage of the second and third options is that the process of pattern extraction and the process of relating the patterns to the ``goodness" of the outcomes are conducted separately.
%%%

%%%
Unlike unsupervised mining, a mining method that directly extracts patterns that are associated with good or bad outcomes is called \emph{supervised mining}.
Roughly speaking, by using supervised mining, we can directly find patterns that have different frequencies depending on the outcomes, thus we can find more direct effects on the outcomes than by simply combining unsupervised mining, as described above.
%%%

\subsection*{Related Work}
\subsubsection*{Sequential pattern mining}
%%%
Sequential pattern mining \cite{agrawal1995mining} involves discovering frequent subsequences as patterns from a database that consists of ordered event sequences, with or without strict notions of time \cite{mabroukeh2010taxonomy}.
Originally applied for the analysis of biological sequences \cite{wang2004scalable, ho2005sequential, exarchos2008mining, hsu2007identification}, sequential pattern mining techniques have also been applied to various other domains including XML document classification \cite{garboni2005sequential}, keyword and key-phrase extraction \cite{feng2011keyword, xie2014document, xie2017efficient}, as well as next item/activity prediction and recommendation systems \cite{yap2012effective,salehi2014personalized,ceci2014completion,wright2015use,tsai2015location,tarus2018hybrid}.
For an overview of the field of sequential pattern mining, we refer the reader to \cite{fournier2017survey}.
%%%

%%%
One of the first sequential pattern mining algorithms was GSP \cite{srikant1996mining}, which was based on earlier work in which the A-priori algorithm was proposed by the same authors \cite{agarwal1994fast}.
SPADE \cite{zaki2001spade}, SPAM \cite{ayres2002sequential}, and the pattern-growth algorithm PrefixSpan \cite{pei2004mining} were proposed to address some limitations that were identified with the GSP algorithm.
PrefixSpan is known as a pattern-growth algorithm, since its grows a tree which extends from a singleton (set with a single event) and adds more events in descendent nodes. 
More recently, CM-SPAM and CM-SPADE \cite{fournier2014fast} as well as Fast \cite{salvemini2011fast} have been proposed to provide further improvements in computational efficiency and therefore speed.
It should be noted that these frequently applied sequential mining algorithms listed above are unsupervised, i.e., are applied to unlabelled sequence data.
%%%

\begin{figure}[!h]
    \centering
	\includegraphics[width=0.4\textwidth]{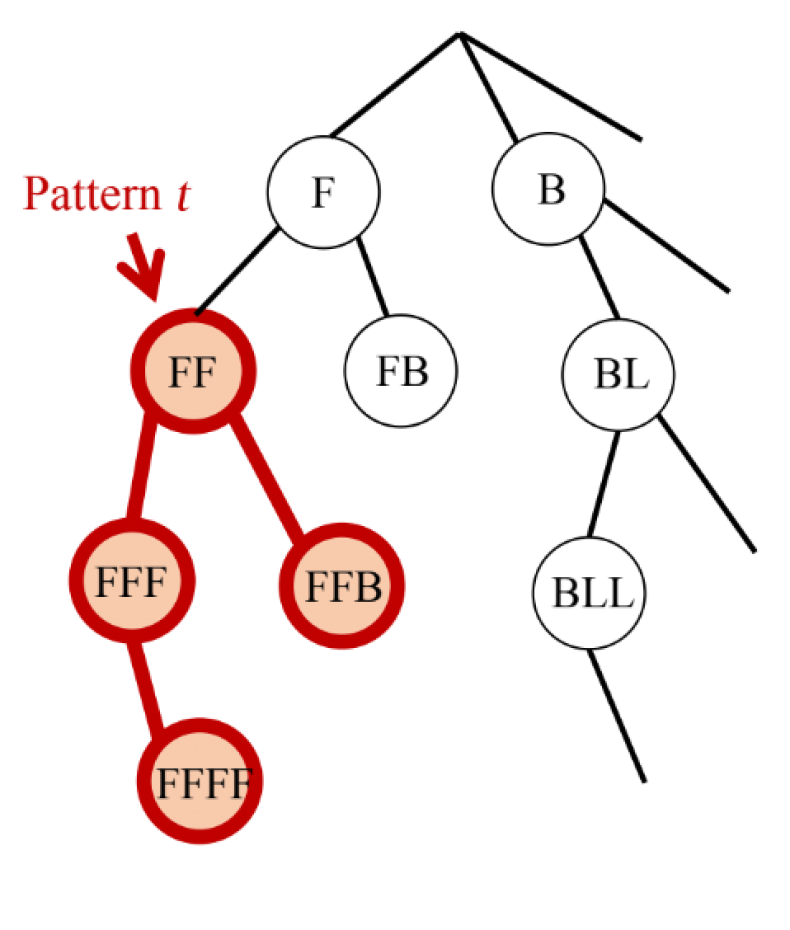}
	\caption{{\bf SPP pruning}. One of the mechanisms within SPP identifies and deletes patterns that do not contribute to the model before performing the optimization. For example, if pattern \textit{t} does not satisfy the SPP pruning criterion specified in \cite{sakuma2019efficient}, the sub-tree below pattern pattern \textit{t} is deleted.}
	\label{fig-spp-tree-pruning}
\end{figure}

%%%
Safe pattern pruning (SPP) was proposed by \cite{nakagawa2016safe,sakuma2019efficient}, and combines a convex optimisation technique called safe screening \cite{ghaoui2010safe} with sequential pattern mining.
SPP is supervised and is applied to labelled data, i.e., to datasets consisting of labelled sequences.
SPP uses PrefixSpan as a building block to grow the initial pattern tree, which is then pruned according to a particular criterion, which prunes the tree structure among all possible patterns in a database, grown by PrefixSpan, in such a way that if a node corresponding to a particular pattern is pruned, it is guaranteed that all patterns corresponding to its descendant nodes are not required for the predictive model (Fig~\ref{fig-spp-tree-pruning}).
%%%
%%%

%
All of the possible pruned patterns in the database are then multiplied by weights in the form of a linear model, and these weights are solved for by solving an optimization problem, however, prior to solving, safe screening is used to eliminate weights that will not be discriminative (i.e., will have values of zero) at the optimal solution.
SPP has been applied to datasets consisting of animal trajectories \cite{sakuma2019efficient}; however, compared with animal trajectories, sports data often contains a greater diversity of events. %CITATION?
%%%
\subsubsection*{Application of sequential pattern mining techniques in sport}
%%%
Unsupervised sequential pattern mining techniques have been applied to data from sport, focusing primarily on the identification, interpretation and visualization of sequential patterns.
Table \ref{tbl-sport-sequential pattern mining} summarizes previous studies that have applied sequential pattern mining techniques to datasets in sport.
CM-SPAM has been applied in order to conduct technical tactical analysis in judo \cite{la2017ontology}.
Sequential data, obtained using trackers, has been used to test for significant trends and interesting sequential patterns in the context of the training of a single cyclist over an extended period of time \cite{hrovat2015interestingness}.
Decroos et al. \cite{decroos2018automatic} combined clustering and CM-SPADE to data from soccer, using a five-step approach, which is presented in Table \ref{tbl-sport-sequential pattern mining}.
Their ranking function allowed the user, e.g., a coach, to assign higher weights to events that are of higher relevance, such as shots and crosses, compared to normal passes, which are very frequent but not necessarily relevant.

\begin{table}[!ht]
	\begin{adjustwidth}{-2.25in}{0in} % Comment out/remove adjustwidth environment if table fits in text column.
	\centering
	\caption{
		{\bf Prior studies that have applied sequential pattern mining techniques in sport.}}
\begin{tabular}{|p{0.08\linewidth}|l|p{0.09\linewidth}|l|p{0.3\linewidth}|p{0.2\linewidth}|}
\thickhline
\textbf{Study} &
  \textbf{Sport} &
  \textbf{Model Used} &
  \textbf{Model Type} &
  \textbf{Summary of Approach} &
  \textbf{Evaluation Metrics} \\ \thickhline
  Hrovat (2015) &
  Cycling &
  SPADE &
  Unsupervised &
  Applied the sequential pattern mining   algorithm SPADE to identify frequent sequential patterns, calculated interestingness   measures (p-values) for these frequent patterns, and visualized these patterns for   increasing/decreasing daily and duration trends &
  Support, permutation test p-values \\ \hline
La Puma \& Giorno (2017) &
  Judo &
  CM-SPAM &
  Unsupervised &
  Identified patterns using sequential pattern mining for the tactical analysis of judo techniques &
  Support \\ \hline

Decroos (2018) &
  Soccer &
  CM-SPADE &
  Unsupervised &
  clustered phases based on   spatio-temporal components, ranked these clusters, mined the clusters to identify   frequent sequential patterns, used a ranking function (a weighted support   function) - in which a coach can assign higher weights to more relevant   events - to score obtained patterns, interpreted the obtained patterns &
  Support (weighted by user's   judgement weighting of the relevance of events), and identified the top-ranked frequent   sequences in the clusters \\ \thickhline
\end{tabular}%
\begin{flushleft}
\end{flushleft}
\label{tbl-sport-sequential pattern mining}
\end{adjustwidth}

\end{table}

%%%

\subsubsection*{Analysis of sequences in rugby union}
%%%
In the sport of rugby union (hereafter referred to simply as rugby) specifically, some previous studies have analyzed matches at the sequence level by analyzing the duration of sequences.
For example, the duration of the sequences of plays leading to tries at the 1995 Rugby World Cup (RWC) were studied by \cite{carter20011995}. 
In a study of the 2003 RWC, \cite{van2006movement} found that teams that were able to create movements that lasted longer than 80 seconds were more successful.
More recently, \cite{coughlan2019they} applied K-modes cluster analysis using sequences of play in rugby, and found that scrums, line-outs and kick receipts were common approaches that led to tries being scored in the 2018 Super Rugby season.
%%%
Recently, \cite{watson2020integrating} used convolutional and recurrent neural networks to predict the outcomes (territory gain, retaining possession, scoring a try, and conceding/being awarded a penalty) of sequences of play, based on event order and their on-field locations.
%%%

\subsection*{Motivation and Contributions}
%%%
In this study, we apply SPP, a supervised sequential pattern mining model, to data consisting of event sequences from all of the matches played by a professional rugby union team in their 2018 Japan Top League season.
The present study is motivated by the fact that, although sequential pattern mining techniques have been applied to sport, only unsupervised models appear to have been used to date. 
In addition, no form of sequential pattern mining technique, unsupervised or supervised, appears to have been applied to the analysis of sequences of play in the sport of rugby union.
%%%

%%%
As a basis for comparison, we also compare the SPP-obtained subsequences with those obtained by well-known unsupervised sequential pattern mining algorithms (PrefixSpan, GSP, Fast, CM-SPADE and CM-SPAM) when they are applied to subsets of the original labelled data, split on the label.
%%%

%%%
The main contributions of this study are in the comparison of the usefulness of supervised and unsupervised sequential pattern mining models that are applied to event sequence data in sport, the application of a supervised sequential pattern mining model to event sequence data in sport, and the application of an sequential pattern mining model for the analysis of sequences of play in rugby.
%%%

\subsection*{Notation}

%%%
The number of unique event symbols is denoted as $m$ and the set of those event symbols is denoted as $\cS := \{s_1, \ldots, s_m\}$. 
In this paper, we refer to sequences and subsequences as \emph{passages} of play and \emph{patterns} of play (or simply \emph{patterns}), respectively.
Let $n$ denote the number of sequences in the dataset ($n$=490 in our dataset).
Sequences with the labels 1 and -1 are denoted as $\cG_{+}, \cG_{-} \subseteq [n]$ and are of size $n_+ := |\cG_+|, n_- := |\cG_-|$, respectively. 
The dataset for building the SPP model is
\begin{align*}
	\{(\bm g_i, y_i)\}_{i \in [n]},
\end{align*}
where
$\bm g_i$
represents the $i$-th sequence/passage of play. Each sequence $\bm g_i$ takes a label from
$y_i \in \{\pm 1\}$
and can be written as 
\begin{align*}
	\bm g_i = \langle g_{i1}, g_{i2}, \ldots, g_{iT(i)} \rangle, i \in [n], 
\end{align*}
where
$g_{it}$
is the $t$-th symbol of the $i$-th sequence, which takes one of the event symbols in $\cS$,
and $T(i)$ indicates the length of the $i$-th sequence, i.e., the number of events in this particular sequence.
Patterns of play are denoted as $\bm q_1, \bm q_2, \ldots$, each of which is also a sequence of event symbols:
\begin{align*}
	\bm q_j = \langle q_{j1}, q_{j2}, \ldots, q_{jL(j)} \rangle, j = 1, 2, \ldots, 
\end{align*}
where $L(j)$ is the length of pattern $\bm q_j$ for $j = 1, 2, \ldots$. 
The relationship whereby sequence $\bm g_i$ contains subsequence $\bm q_j$ is represented as $\bm q_j \sqsubseteq \bm g_i$. 
The set of all possible patterns contained in any sequence
$\{\bm g_i\}_{i \in [n]}$
is denoted as
$\cQ = \{ \bq_i \}_{i \in [d]}$, where $d$ is the number of possible patterns (large in general).

\section*{Materials and Methods}
\label{sec-materials-methods}
\subsection*{Data}
\begin{figure}[!h]
 	\centering
	\includegraphics[width=1\textwidth]{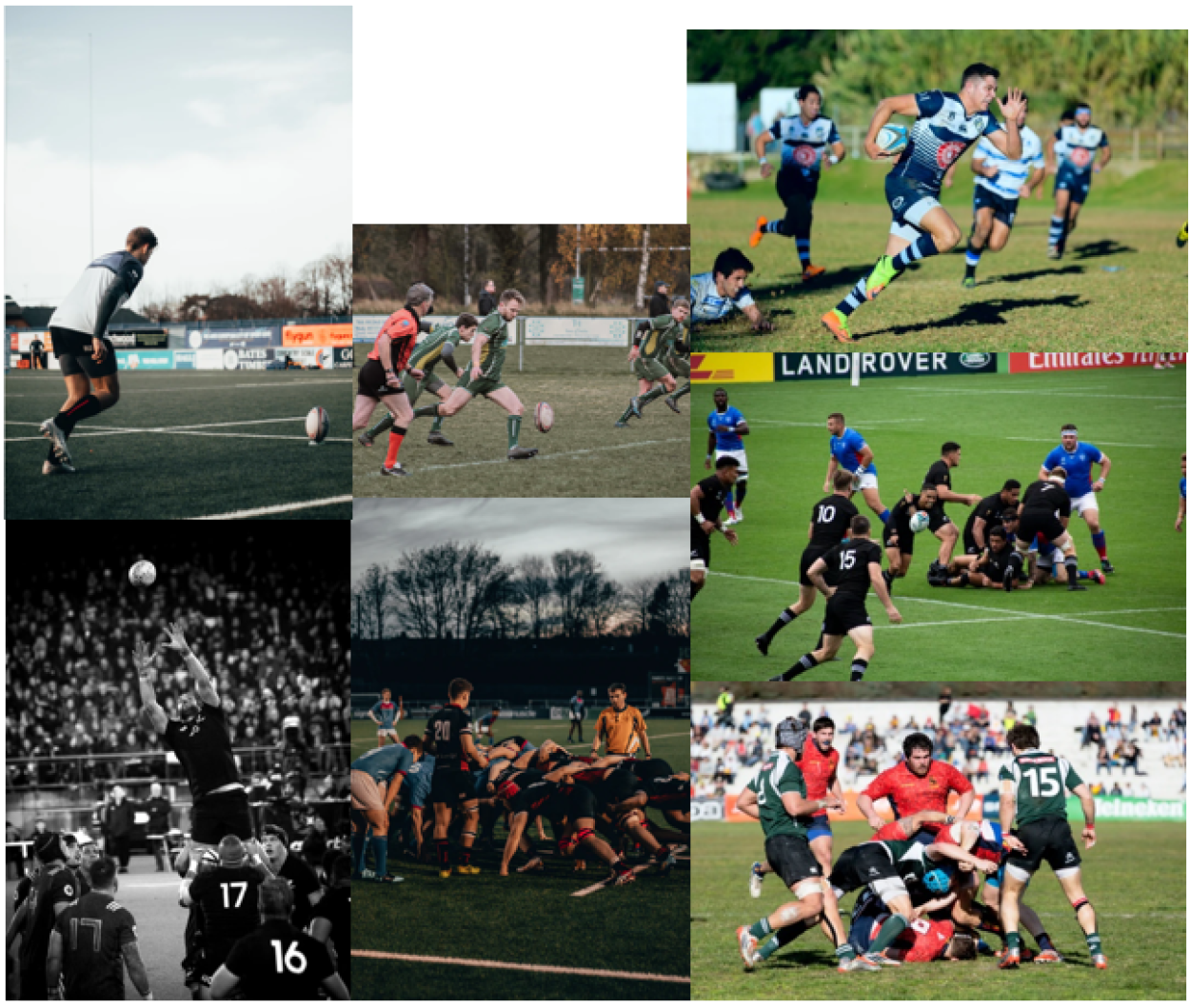}
	\caption{{\bf Key events in rugby matches.} The photographs used as the original images are listed in parentheses. All of them are licensed under the unsplash.com license (\url{https://unsplash.com/license}). Top left: Kick at goal (\url{https://unsplash.com/photos/xJSPP3H8XTQ});
Bottom left: Lineout (\url{https://unsplash.com/photos/CTEvFbFpVC8});
Center top: Kick restart/Kick-off (\url{https://unsplash.com/photos/OMdge7F2FyA});
Center bottom: Scrum (\url{https://unsplash.com/photos/y5H3\_7OobJw});
Top Right: Linebreak (\url{https://unsplash.com/photos/XAlKHW9ierw)};
Middle Right: Beginning of a phase (\url{https://unsplash.com/photos/fqrzserMsX4)};
Bottom Right: Breakdown (\url{https://unsplash.com/photos/WByu11skzSc)}}
	\label{fig-rugby-events}
\end{figure}

%%%
We obtained XML data generated from video tagged in Hudl Sportscode (\url{https://www.hudl.com/products/sportscode}) by the performance analyst of one of the teams in the Japan Top League competition (not named for reasons of confidentiality). 
Written consent was obtained to use the data for research purposes.
Seasons are comprised of a number of matches, matches are made up of sequences of play, which are, in turn, comprised of events.
Our dataset consisted of all of this particular team's matches in their 2018 season against each of the opposition teams they faced. 
These matches consist of passages of play (i.e., sequences of events), however, each match in the original dataset were each one long sequence.
One approach is to label as sequences with win/loss outcomes, however, in our initial trials, this did not produce interesting results since it is obvious that sequences containing a greater number of scoring events will be within match-sequences labelled with wins.
Therefore, we generated a dataset that is of greater granularity by defining rules that delimit matches into sequences representing passages of play (we outline these in the following subsection).
The 24 unique events (12 unique events for the team and opposition teams), in our data are listed in Table \ref{tbl-original-events}, and some are also depicted in Fig \ref{fig-rugby-events}.
The XML data also contained a more granular level of data than these 24 events represent (i.e., with more detailed events---in other words, a larger number of events); however, in order to reduce computational complexity, the higher level of the data was considered.
%%%

\begin{table}[!ht]
	\begin{adjustwidth}{-2.25in}{0in} % Comment out/remove adjustwidth environment if table fits in text column.
	\centering
	\caption{
		{\bf Unique events in the original XML data.} Events prefixed by "O-" are performed by the opposition team, while those that are not a performed by the team.}
	\begin{tabular}{|l|l|l|}
		\thickhline
		{\bf event ID} & {\bf event} & {\bf event description} \\ \thickhline
		1 & Restart Receptions & Team receives a kick restart made by the opposition team\\ \hline
		2 & Phase & Period between breakdowns (team in possession of the ball) \\ \hline
		3 & Breakdown & Team player is tackled, resulting in a ruck \\ \hline
		4 & Kick in Play & Kick within the field of play (rather than to touch) made by the team\\ \hline
		5 & Penalty Conceded & Team gives away a penalty, opposition may re-gain possession \\ \hline
		6 & Kick at Goal & Team attempts kick at goal\\ \hline
		7 & Quick Tap & Quick restart of play by the team following a free kick awarded to them\\ \hline
		8 & Lineout & Ball is thrown in by the team \\ \hline
		9 & Error & Mistake made by the team, e.g., lost possession, forward pass, etc. \\ \hline
		10 & Scrum & Set piece in which the forwards attempt to push the opposing team off the ball \\ \hline
		11 & Try Scored & Team places the ball down over opposition team's line (five points) \\ \hline
		12 & Line Breaks & Team breaches the opposition team's defensive line \\ \hline
		13 & O-Restart Receptions & Opposition team receives a kick restart made by the team\\ \hline
		14 & O-Phase & Period between breakdowns (opposition team in possession of the ball)\\ \hline
		15 & O-Breakdown & Opposition player is tackled, resulting in a ruck\\ \hline
		16 & O-Kick in Play & Kick within the field of play (rather than to touch) made by the opposition team\\ \hline
		17 & O-Penalty Conceded & Opposition team gives away a penalty, team may re-gain possession \\ \hline
		18 & O-Kick at Goal & Opposition team attempts kick at goal\\ \hline
		19 & O-Quick Tap & Quick restart of play by the opposition team following a free kick awarded to them\\ \hline
		20 & O-Lineout & Ball is thrown in by the opposition team\\ \hline
		21 & O-Error & Mistake made by the opposition team, e.g., lost possession, forward pass, etc. \\ \hline
		22 & O-Scrum & Set piece in which the forwards attempt to push the team off the ball \\ \hline
		23 & O-Try Scored & Opposition team places the ball down over the team's line (five points)  \\ \hline
		24 & O-Line Breaks & Opposition team breaches the team's defensive line \\ \thickhline
		%label & - & Points Scored & O-Points Scored\\ \hline
		%& $n$=490 & $n_+$=86, $n_-$=404\\ \thickhline
	\end{tabular}
	\begin{flushleft} 
	\end{flushleft}
	\label{tbl-original-events}
	\end{adjustwidth}
\end{table}

\subsection*{Methods}
%%%
\subsubsection*{Delimiting matches into sequences}
\begin{figure}[!h]
    \begin{adjustwidth}{-2.25in}{0in}
	\includegraphics[width=1.5\textwidth]{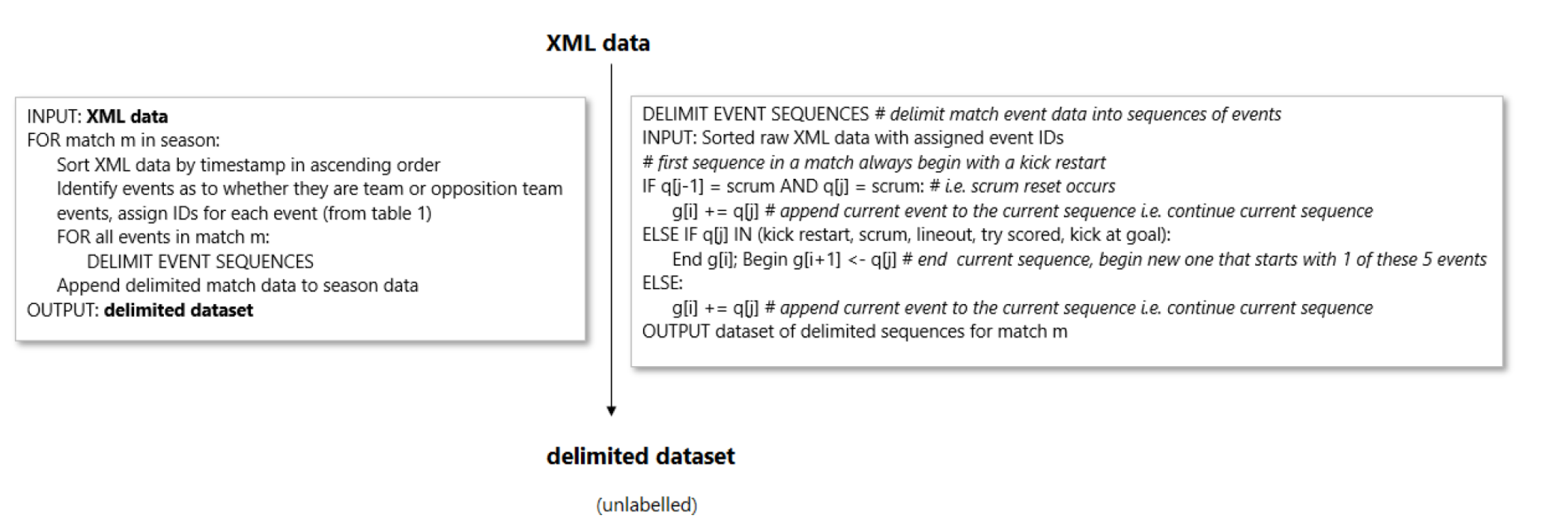}
	\caption{{\bf Illustration of the procedure to delimit the raw XML data into labelled sequences of events.}}
	\label{fig-process-flow-1}
	\begin{flushleft}
	\end{flushleft}
	\end{adjustwidth}
\end{figure}

Our dataset was converted into labelled event sequences by delimiting each match into passages of play (Fig~\ref{fig-process-flow-1})
The rules to delimit matches into sequences of events (passages of play), should ideally begin and end at logical points in the match, e.g., when certain events occur, when play stops or when possession changes (e.g., \cite{liu2018using}), and should result in sequences which are neither overly long nor overly short.
In this study, a passage of play was defined to start with either a kick restart, scrum, or lineout, which are events that result in play temporarily stopping and therefore represent natural delimiters for our dataset.
When there is a kick restart, scrum (except for a scrum reset where a scrum follows another scrum), or lineout, this event becomes the first event in a new event sequence; otherwise, if a try is scored or a kick at goal occurs, a new passage of play also begins.
Applying these rules (also shown in Fig~\ref{fig-process-flow-1}) resulted in delimited dataset consisted of 490 sequences.
Each of these sequences were made up of events from Table \ref{tbl-original-events}.
At this stage, the delimited dataset is unlabelled, with the scoring events (try scored, kick at goal) for the team and opposition teams contained in the sequences.

\subsubsection*{Experimental dataset creation and comparative approach}
%%%
The delimited dataset described was then divided into two datasets.
In the first, which we call the scoring dataset, we consider the case where the sequences are from the team's scoring perspective. 
In this dataset, the label $y_i = +1$ represents points being scored or attempted. 
Note that while a try scored was certain in terms of points being scored, a kick at goal (depicted in the top-left of Fig \ref{fig-rugby-events}) is not always successful.
In our data, only the kick at goal being attempted (event id 6) was available---not whether the goal was actually successful or not. 
However, since it is more important to be able to identify points-scoring opportunities than whether or not the kick was ultimately successful (which is determined by the accuracy of the goal kicker), we assume that 100\% of kicks at goal resulted in points being scored. 
In the scoring dataset, the label $y_i = +1$ was assigned assigned to the sequences from the original delimited dataset if a try was scored or a kick at goal was made by the team in sequence $i$.
If there was no try scored and no kick at goal made by the team in sequence $i$, the label $y_i = -1$ was assigned.
Then, since the label now identifies scoring/not scoring, the events that relate to the team scoring---Try scored (event ID = 11) and Kick at goal (event ID = 6)---were removed from the event sequences.
%%%

%%%
In the second, which we call the conceding dataset, we consider the case where the sequences are from the team's conceding perspective, or equivalently, from the opposition teams' scoring perspective. 
In the conceding dataset, the label $y_i = +1$ was assigned to the sequences from the original delimited dataset if a try was scored or a kick at goal was made by the \textit{opposition} team in sequence $i$.
If there was no try scored and no kick at goal made by the \textit{opposition} team in sequence $i$, the label $y_i = -1$ was assigned.
The list of events for the original delimited, scoring and conceding datasets are presented in Table \ref{table1}.
Then, since the label now identifies scoring/not scoring, the events that relate to the \textit{opposition} team scoring---Try scored (event ID = 11) and Kick at goal (event ID = 6)---were removed from the event sequences.
%%%

\begin{figure}[!h]
    \begin{adjustwidth}{-2.25in}{0in}
	\includegraphics[width=1.5\textwidth]{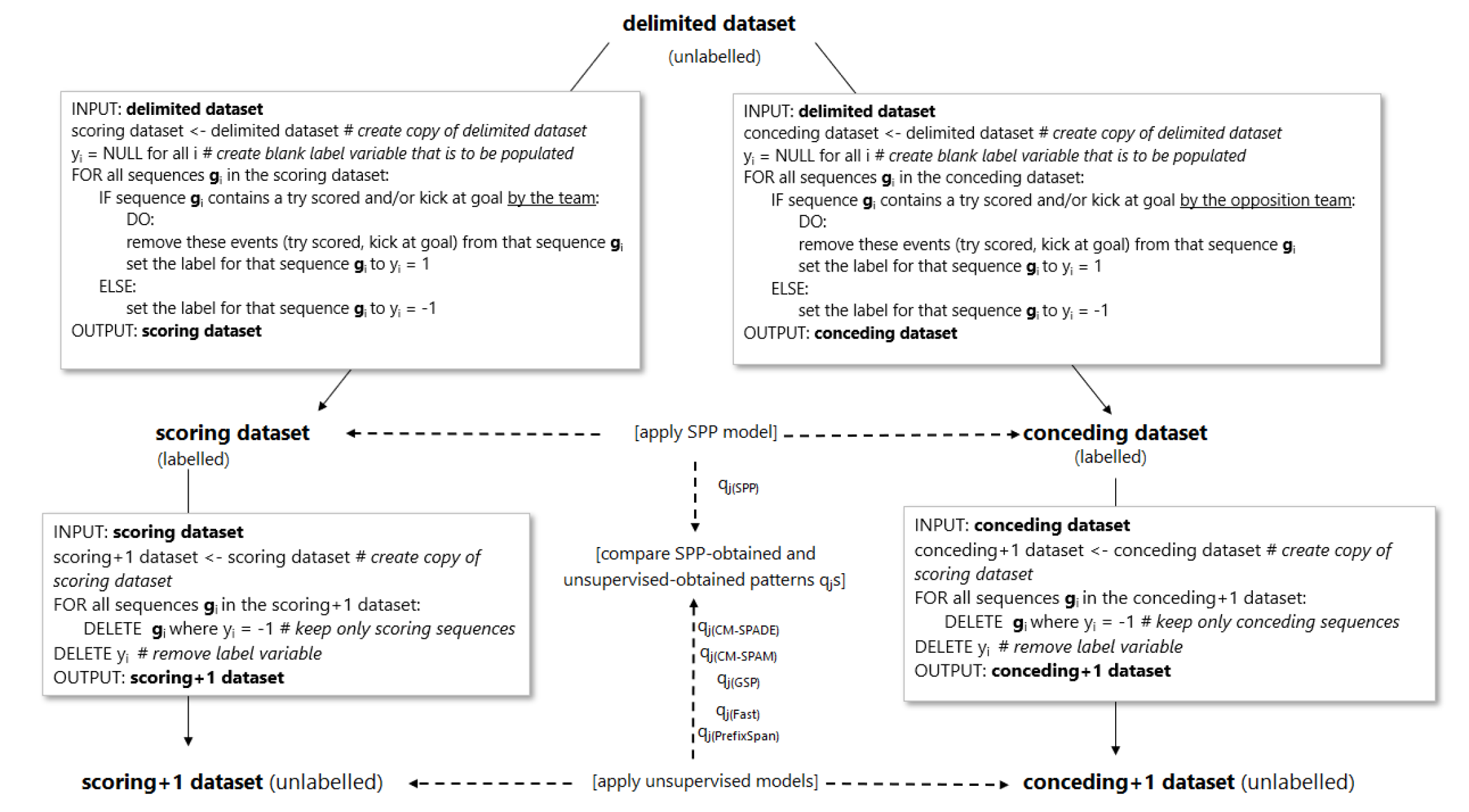}
	\caption{{\bf Illustration of dataset creation and experimental approach.}
		Illustration of the procedures to create the datasets from the original delimited dataset to be used in the experiments and to compare the unsupervised and supervised sequential pattern mining models.}
	\label{fig-process-flow-2}
	\begin{flushleft}
	\end{flushleft}
	\end{adjustwidth}
\end{figure}

%%%
The process applied to create the scoring and conceding datasets from the original delimited dataset is shown in the upper half of Fig~\ref{fig-process-flow-2}.

\begin{table}[!ht]
	%	\begin{adjustwidth}{-2.25in}{0in} % Comment out/remove adjustwidth environment if table fits in text column.
	\centering
	\caption{
		{\bf Event lists for the original, scoring and conceding datasets.}}
	\begin{tabular}{|l|l|l|l|}
		\thickhline
		{\bf event ID} & {\bf original} & {\bf scoring} & {\bf conceding} \\ \thickhline
		1 & Restart Receptions & Restart Receptions & Restart Receptions \\ \hline
		2 & Phase & Phase & Phase \\ \hline
		3 & Breakdown & Breakdown & Breakdown \\ \hline
		4 & Kick in Play & Kick in Play & Kick in Play \\ \hline
		5 & Penalty Conceded & Penalty Conceded & Penalty Conceded \\ \hline
		6 & Kick at Goal &  & Kick at Goal \\ \hline
		7 & Quick Tap & Quick Tap & Quick Tap \\ \hline
		8 & Lineout & Lineout & Lineout \\ \hline
		9 & Error & Error & Error \\ \hline
		10 & Scrum & Scrum & Scrum \\ \hline
		11 & Try Scored &  & Try Scored \\ \hline
		12 & Line Breaks & Line Breaks & Line Breaks \\ \hline
		13 & O-Restart Receptions & O-Restart Receptions & O-Restart Receptions \\ \hline
		14 & O-Phase & O-Phase & O-Phase \\ \hline
		15 & O-Breakdown & O-Breakdown & O-Breakdown \\ \hline
		16 & O-Kick in Play & O-Kick in Play & O-Kick in Play \\ \hline
		17 & O-Penalty Conceded & O-Penalty Conceded & O-Penalty Conceded \\ \hline
		18 & O-Kick at Goal & O-Kick at Goal &  \\ \hline
		19 & O-Quick Tap & O-Quick Tap & O-Quick Tap \\ \hline
		20 & O-Lineout & O-Lineout & O-Lineout \\ \hline
		21 & O-Error & O-Error & O-Error \\ \hline
		22 & O-Scrum & O-Scrum & O-Scrum \\ \hline
		23 & O-Try Scored & O-Try Scored &  \\ \hline
		24 & O-Line Breaks & O-Line Breaks & O-Line Breaks \\ \thickhline
		label & - & Points Scored & O-Points Scored\\ \hline
		& $n$=490 & $n_+$=86, $n_-$=404 & $n_+$=44, $n_-$=446\\ \thickhline
	\end{tabular}
	\begin{flushleft} 
	%From the team's scoring perspective, the team's points scored (event 6 and 11) determine the class label for the sequences, while opposition points scored (event 18 and 23) are treated as events in the sequences. From the team's conceding perspective, opposition points scored (events 18 and 23) determine the class label for the sequences, while opposition points scored (events 6 and 11) are treated as events in the sequences. Events prefixed by ``O-" indicate that the events relate to the opposition team; those that are not are events that relate to the team. 
		%Recall that the scoring dataset label Points Scored = 1 if Try Scored and/or Kick at Goal are in the sequence, 0 otherwise. Similarly, the conceding dataset label O-Points Scored = 1 if O-Try Scored and/or O-Kick at Goal are in the sequence, 0 otherwise.
	\end{flushleft}
	\label{table1}
	%	\end{adjustwidth}
\end{table}

%%%
The SPP algorithm (software is available at \url{https://github.com/takeuchi-lab/SafePatternPruning}) was applied to the scoring and conceding datasets.
%%%

%%%
As a basis for comparison, we compare the obtained subsequences ($q_{j}$s) from SPP with those obtained by the unsupervised algorithms: PrefixSpan, CM-SPAM, CM-SPADE, GSP and Fast. 
The SPMF pattern mining package \cite{fournier2014sequential pattern miningf} (v2.42c) was used for the application of the five unsupervised sequential pattern mining algorithms to our dataset.
Since the unsupervised models use unlabelled data, while support values of the patterns of play can be obtained, we cannot obtain weights for the patterns.
For a more fair comparison between the unsupervised models and the supervised model, SPP, we assume prior knowledge of the sequence labels to apply the unsupervised models.
Thus, the unsupervised models were applied to the dataset, which we call ``scoring+1," containing the sequences where the team actually scored, and to the ``conceding+1" dataset, containing the sequences where the team actually conceded points (i.e., the opposition team scored points).
%%%

%%%
The dataset creation process and comparative approach is presented in Fig~\ref{fig-process-flow-2}.
%%%

\subsubsection*{Obtaining pattern weights with safe pattern pruning}
%%%
As mentioned, our data consists of sequences comprised of events from Table \ref{tbl-original-events}, which are labelled with an outcome: either +1 or -1, e.g. \newline
\texttt{-1    22    22    17}												
\newline
\texttt{1    8    11    2    6}													\newline													
\texttt{-1	1	2	3	2	9}														\newline												
\texttt{-1	20	21}																	\newline														
\texttt{-1	10	10	2	3	2	3	2	3	2	3	2	17}							\newline														
\texttt{1	8	11	2	6}															\newline													
\texttt{-1	1	2	3	2	3	9}													\newline													
\texttt{-1	22	16	2}																\newline													
\texttt{-1	13	14	16}
\newline
...
\newline
We are interested in using SPP to identify subsequences of events that discriminate between outcome +1 and outcome -1.
For instance, in the dataset above, it would seem that subsequence [2,3,2] is potentially a discriminative pattern, since it appears in three sequences that are labeled with -1 and none that are labeled with 1, while [11,2,6] is also potentially a discriminative pattern since it appears in two sequences with label 1 and none with -1.
SPP involves taking linear combinations of the subsequences with weights, e.g., $w_1$[2,3,2] + $w_2$[11,2,6]..., for each sequence, and then to use an optimization model to calculate these weights.
Discriminative patterns have positive absolute values at the optimal solution.
%%%

%%%
A classifier based on a sparse linear combinations of patterns can be written as
\begin{align}
	f(\bg_i ; \cQ) = \sum_{\bq_j \in \cQ} w_j I(\bq_j \ssse \bg_i) + b,
	\label{eq:linear-weighted-subsequence-model}
\end{align}
where $I(\cdot)$ is an indicator function that takes the value 1 if sequence $\bg_i$ contains subsequence $\bg_i$ and 0 other otherwise; and $w_j \in \RR$ and $b \in \RR$ are parameters of the linear model, which are estimated by solving the following minimisation problem (as well as its dual maximization problem; see \cite{sakuma2019efficient} for details of the pruning criterion):
\begin{align}
	\min_{\bw, b} \sum_{i \in [n]} \ell(y_i, f(\bg_i ; \cQ)) + \lambda \| \bw \|_1,
	\label{eq:LWSM-optimization}
\end{align}
where
$\bw = [w_1, \ldots, w_d]^\top$ is a vector of weights, $\ell$ is a loss function and $\lambda > 0$ is a regularization parameter that can be tuned by cross-validation. 
Note that, due to the permutations in terms of the number of potential patterns of play, the size of $\cQ$ is quite large in general.
The goal of SPP is to reduce the size of $\cQ$ by removing unnecessary patterns from the entire pattern-tree that was grown by PrefixSpan according to the SPP pruning criterion \cite{sakuma2019efficient}.
The minimization problem \eqref{eq:linear-weighted-subsequence-model} was, in the present study, solved with an L1-regularised L2-Support Vector Machine (the default option -u 1 in the S3P classifier command line options \url{https://github.com/takeuchi-lab/S3P-classifier}), with 10-times-10 cross-validation used to tune the regularization parameter lambda (options -c 1 -M 1 in the S3P classifier command line options).
The maximum pattern length parameter  (option -L in the S3P classifier command line options) was set to 20.
The feature vector $\bx_i = [x_{i1}, x_{i2},\ldots, x_{id}]$ is defined for the $i$th sequence $\bg_i$ as 
\begin{align}
	x_{ij} = I(\bq_j \ssse \bg_i), \hspace{5em} j = 1, \ldots, |\cQ|.
\end{align}
In other words, the feature vectors $\bx_i = [I(\bq_1 \ssse \bg_i), I(\bq_2 \ssse \bg_i),\ldots, I(\bq_d \ssse \bg_i)]$ are binary variables that take the respective values 1 or 0 based on whether or not subsequence $\bq_j$ is contained within sequence $\bg_i$.
The squared hinge-loss function $\ell(y,f(\bx_i)) = \max\{0, 1 - y f(\bx_i) \}^2$ is used for a two-class problem like ours, in which case the optimization problem (\ref{eq:LWSM-optimization}) becomes:
\begin{align}
	\min_{\bw, b} \sum_{i \in [n]} \max \left\{ 0, 1 - y_i (\bw^\top \bx_i + b) \right\}^2 + \lambda \| \bw \|_1,
	\label{eq:L2-SVM}
\end{align}
Discriminative patterns are those that have positive weights (in absolute terms) in the optimal solution to (\ref{eq:L2-SVM}) (in SPP, some weights are removed prior to solving the optimization problem by using safe screening---see \nameref{S2_Appendix} for more details).
%

%%%
In this study, in order to exclude patterns that may have occurred merely by chance, the obtained patterns ($q_{j}$s) for all datasets with support of less than five were removed.
In the case of the patterns obtained by the unsupervised model, the top five patterns with the largest support values were recorded.
In the case of the SPP-obtained patterns, the top five patterns with the largest positive $w_{j}$ values were recorded.
In addition, we restricted our analysis to patterns of play that had the highest positive weights. 
For the scoring dataset, this means the patterns that had a positive contribution to the team scoring. 
For the conceding dataset, this means the patterns that had a positive contribution to opposition teams scoring. 
In other words, for the sake of brevity, we did not consider the patterns that had the highest contribution to ``not scoring'' and ``not conceding.''
The obtained results are presented in the following section.
%%%

\section*{Results}
\label{sec-results}
\subsection*{Analysis of sequence lengths}
\begin{figure}[!h]
	\includegraphics[width=1\textwidth]{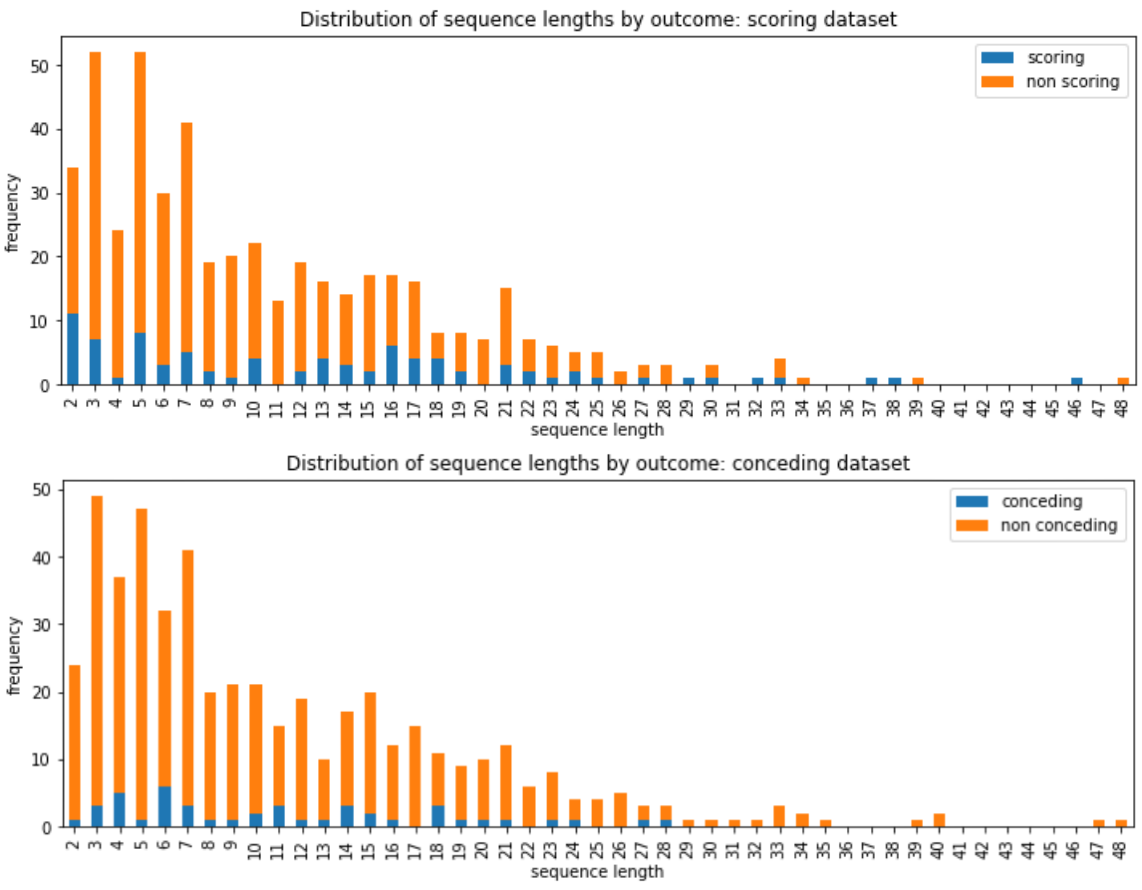}
	\caption{{\bf Sequence length distributions.}
		Distribution of sequence lengths by points-scoring outcome for the scoring and conceding datasets. Sequence length is defined as the number of events in each sequence (excluding the outcome label).}
	\label{fig-seq-len-dists}
\end{figure}

%%%
There were an average of 10.6 events in each sequence in the scoring dataset, and 10.8 events in the conceding dataset. 
The shortest sequence contained two events, and the longest contained 48 events (Table \ref{table2}).
The slight differences in mean sequence lengths between the scoring and conceding datasets is a result of the removal of the try and kick at goal events from the sequences in order to create the sequence outcome label (as mentioned in the Materials and Methods section above). 
The sequence length distributions are positively skewed and non-normal (Fig~\ref{fig-seq-len-dists}), which was confirmed by Shapiro-Wilk tests.
By comparing these distributions, it is clear that the number of sequences in which points were scored was higher in the scoring dataset than the conceding dataset, which is reflective of the strength of the team in the 2018 season.
From the team's scoring perspective, 86 out of the 490 passages of play (18\%) resulted in points being scored by the team, while from the team's conceding perspective, 44 out of the 490 passages of play (9\%) resulted in points conceded. 
The sequences in which the team scored points were slightly longer, containing 12.8 events on average compared to those where the team didn't score, which contained 10.2 events, on average.
The sequences in which the team conceded points contained 11.2 events on average, while those where the team didn't concede points contained 10.8 events, on average.
%%%

\begin{table}[]
	%\begin{adjustwidth}{-2.25in}{0in} % Comment out/remove adjustwidth environment if table fits in text column.
		\centering
		\caption{
			{\bf Descriptive statistics for the scoring and conceding datasets}}
		\begin{tabular}{|l|l|l|}
			\thickhline
			{\bf } & {\bf scoring} & {\bf conceding} \\ \thickhline
			Mean & 10.6 & 10.8 \\ \hline
			Standard deviation & 7.8 & 7.9\\ \hline
			Minimum & 2 & 2 \\ \hline
			25th percentile & 5 & 5\\ \hline
			Median & 8 & 8\\ \hline
			75th percentile & 15 & 15\\ \hline
			Maximum & 48 & 48\\ \hline
			Skewness & 1.3 & 1.4\\ \thickhline
	\end{tabular}
\begin{flushleft} 
\end{flushleft}
\label{table2}
%\end{adjustwidth}
\end{table}

\subsection*{Identification of important patterns of play using SPP}
\label{subsec-important-patterns}
%%%
SPP initially obtained 93 patterns when applied to the scoring dataset, of which 75 had support of 5 or higher. 
Out of these 75 patterns of play, 38 had a positive weight ($w_j > 0$). 
The 75 patterns with minimum support of 5 contained an average of 4.5 events, and the 38 patterns with positive weights contained an average of 5.4 events. 
The longest obtained pattern in the scoring dataset contained 16 events.
%%%

%%%
Applying SPP to the conceding dataset resulted in a total of 72 patterns, of which 51 had support of 5 or higher.
Out of these 51 patterns of play, 31 had a positive weight ($w_j > 0$).
The 51 patterns with minimum support of 5 contained an average of 3.8 events, and the 31 patterns with positive weights contained an average of 4.4 events.
The longest obtained pattern in the conceding dataset contained 15 events.
%%%

%%%
The five most discriminative patterns between scoring and non-scoring outcomes (i.e., patterns with the highest positive weight contributions) were obtained by applying SPP to the scoring dataset, and are listed along with their weight values and odds ratios in Table \ref{table3}.
In the results tables, the notation [p] x n, denotes that pattern p is repeated n times.
We include the odds ratio (OR) for these patterns (simply the exponential of the weight), which aids in interpretation by providing a value that compares the cases where a sequence contains a particular pattern, and when it does not.
%%%

%%%
The pattern in the scoring dataset with the highest weight value (0.919), which discriminated the most between scoring and non-scoring sequences, was a pattern with a single line break event (event id 12). 
The OR for the linebreak pattern is exp(0.919)=2.506, meaning that the team is 2.5 times more likely to score when a line break occurs in a sequence of play than if a line break is not made in a sequence of play.
Line breaks, which involve breaking through an opposition team's line of defense (see the top-right image in Fig \ref{fig-rugby-events}), advance the attacking team forward and are thus expected to create possible scoring opportunities.
A lineout followed by phase play (8 2) was the second most discriminative pattern between scoring and not scoring, with a weight of 0.808 and an OR of 2.242, indicating that the team is 2.2 times more likely to score when a lineout followed by a phase occurs in a sequence of play than if it does not.
The third most discriminative pattern, 2 3 4 2 3 (w=0.796, OR=2.217), can be interpreted as a kick in play being made by the team and being re-gathered by the team, thus resulting in retained possession.
This indicates that the team is 2.2 times more likely to score when this pattern occurs in a sequence of play than if it does not.
The fourth most discriminative pattern, 2 3 2 3 2 3 2 3 4 (w=0.732, OR=2.079), represents four repeated phase-breakdown plays by the team, followed by the team making a kick in play, which indicates repeated retaining of possession before presumably gaining territory in the form of a kick.
This indicates that the team is 2.1 times more likely to score when this pattern occurs in a sequence of play than if it does not.
The fifth most discriminative pattern, 13 14 15 14 15 16 14 2 3 (w=0.710, OR=2.033), can be interpreted as the opposition team receiving a kick restart made by the team, attempting to exit their own territory via a kick but not finding touch, thus giving the ball back to the team from which they can potentially build phases and launch an attack.
This indicates that the team is twice as likely to score when this pattern occurs in a sequence of play than if it does not.
%%%

\begin{table}[!ht]
	\begin{adjustwidth}{-2.25in}{0in} % Comment out/remove adjustwidth environment if table fits in text column.
		\centering
		\caption{
			{\bf Top five most discriminative SPP-obtained patterns between scoring and non-scoring outcomes.}}
		\begin{tabular}{|l|p{10cm}|l|l|l|}
			\thickhline
			\bf{pattern ($q_{j}$)} & \bf{pattern description} & \bf{support} &  \bf{weight} &  \bf{OR}\\ \thickhline
			12 & linebreak & 77 & 0.919 & 2.506\\ \hline
			8 2 & lineout, phase & 71 & 0.808 & 2.242\\ \hline
			2 3 4 2 3 & phase, breakdown, kick in play, phase, breakdown & 9 & 0.796 & 2.217 \\ \hline
			2 3 2 3 2 3 2 3 4 & {[}phase, breakdown{]}x4, kick in play & 9 & 0.732 & 2.079\\ \hline
			13 14 15 14 15 16 14 2 3 & O-restart received, {[}O-phase, O-breakdown{]}x2, O-kick in play, phase, breakdown & 6  & 0.710 & 2.033\\ \thickhline
		\end{tabular}
		\begin{flushleft} 
		\end{flushleft}
		\label{table3}
	\end{adjustwidth}
\end{table}

%%%
The five most discriminative patterns between conceding and non-conceding outcomes (i.e., patterns with the highest positive weights) were obtained by applying SPP to the conceding dataset, and are listed along with their weight values and odds ratios in Table \ref{table4}.
A linebreak (event ID 24) (w=0.613, OR=1.846) being made by the opposition team was the most discriminative pattern between sequences in which the team conceded and did not concede, or in other words, a linebreak by the opposition team was the pattern that discriminated the most between the group of sequences in which the opposition team scored and the group of sequences in which the opposition team did not score.
The weight magnitude was not as large as for the team scoring from a linebreak against the opposition team (w=0.919 vs. w=0.613), suggesting that the team has strong defence since linebreaks by the opposition team were less likely to result in the opposition team scoring compared to the likelihood of linebreaks made by the team through the opposition defensive line resulting in them scoring.
The OR of 1.8 indicated that the opposition team is 1.8 times more likely to score when they make a linebreak in a sequence of play than if they do not.
The second most discriminative pattern 14 9 15 (w=0.392, OR=1.479) between conceding and non-conceding outcomes can be interpreted as the opposition team being in possession of the ball, the team making some form of error, and the opposition team regaining possession.
The opposition team is 1.5 times more likely to score when this pattern occurs in a sequence of play than if it does not.
The third most discriminative pattern (20) between conceding and non-conceding outcomes was an opposition team lineout (w=0.357, OR=1.428).
The opposition team is 1.4 times more likely to score if they have a lineout in a sequence of play than if they do not.
The fourth (w=0.339, OR=1.403) and fifth (w=0.261, 1.299) most discriminative patterns for the conceding dataset represent repeated phase and breakdown play, with the fifth subsequence, for example, indicating the opposition team making over six repeated consecutive phases and breakdowns, suggesting the retaining of possession and building of pressure by the opposition team.
%%%
\begin{table}[!ht]
	\begin{adjustwidth}{-2.25in}{0in} % Comment out/remove adjustwidth environment if table fits in text column.
%		\centering
		\caption{
			{\bf Top five most discriminative SPP-obtained patterns between conceding and non-conceding outcomes.}}
		\begin{tabular}{|l|p{7cm}|l|l|l|}
			\thickhline
			\bf{event id pattern ($q_{j}$)} & \bf{pattern description} & \bf{support} & \bf{weight} & \bf{OR}\\ \thickhline
			%1,2,3,4,2,5 & restart reception, phase, breakdown, kick in play, phase, penalty conceded & 2 & 0.952 & 2.591\\ \hline
			24 & O-Linebreak & 32 & 0.613 & 1.846\\ \hline
			14 9 15 & O-phase, error, O-breakdown & 10 & 0.392 & 1.479\\ \hline
			20 & O-lineout & 86  & 0.357 & 1.428\\ \hline
			15 15 14 15 & O-breakdown, O-breakdown, O-phase, O-breakdown & 5 & 0.339 & 1.403 \\ \hline
			15 14 15 14 15 14 15 14 15 14 15 14 15 & {[}O-breakdown, O-phase{]}x6, O-breakdown & 16   & 0.261 & 1.299\\ \thickhline
	\end{tabular}
\begin{flushleft} 
\end{flushleft}
\label{table4}
\end{adjustwidth}
\end{table}

\subsection*{Comparison of SPP-obtained patterns to those obtained by unsupervised models}
\label{subsec-comparison-unsupervised}
%%%
Tables \ref{table5} and \ref{table6} show the top five subsequences in terms of their support from the scoring+1 and conceding+1 datasets.
%%%

\begin{table}[!ht]
%\begin{adjustwidth}{-2.25in}{0in}
%		\centering
		\caption{
			{\bf Top five PrefixSpan-obtained patterns of play with the largest support: scoring+1 dataset.}}
\begin{tabular}{|l|l|l|l|l|l|}
\thickhline
\bf{PrefixSpan} & \bf{CM-SPAM} & \bf{CM-SPADE} & \bf{GSP} & \bf{Fast} & \bf{support}\\ \thickhline
2          & 2         & 2          & 2   & 2    & 84\\ \hline
2 3        & 3         & 3          & 3   & 3    & 60\\ \hline
3          & 2 3       & 2 3        & 2 3 & 2 3  & 60\\ \hline
2 2        & 2 2       & 3 2        & 2 2 & 2 2  & 59\\ \hline
2 3 2      & 2 3 2     & 2 2        & 3 2 & 3 2  & 59\\ \thickhline
\end{tabular}
%\begin{flushleft} 
%\end{flushleft}
\label{table5}
%\end{adjustwidth}
\end{table}

\begin{table}[!ht]
%\begin{adjustwidth}{-2.25in}{0in}
%		\centering
		\caption{
			{\bf Top five PrefixSpan-obtained patterns of play with the largest support: conceding+1 dataset.}}
\begin{tabular}{|l|l|l|l|l|l|}
\thickhline
\bf{PrefixSpan} & \bf{CM-SPAM} & \bf{CM-SPADE} & \bf{GSP} & \bf{Fast} & \bf{support}\\ \thickhline
14          & 14         & 14          & 14   & 14    & 39\\ \hline
14 15        & 15         & 15          & 15   & 15    & 33\\ \hline
15          & 14 15       & 14 15        & 14 15 & 14 15  & 33\\ \hline
14 14        & 14 14      & 15 14        & 14 14 & 14 14  & 29\\ \hline
14 15 14      & 14 15 14     & 14 14        & 15 14 & 15 14  & 29\\ \thickhline
\end{tabular}
%\begin{flushleft} 
%\end{flushleft}
\label{table6}
%\end{adjustwidth}
\end{table}

%%%
The obtained results show that common events and patterns were detected with the unsupervised models, i.e., breakdowns and phases. 
Repeated breakdown and phase play is a means retaining possession of the ball and building pressure (see the middle and bottom images on the right-hand side of Fig \ref{fig-rugby-events}).
Longer repeated breakdown and phases plays were also identified by SPP.
However, in the case of the unsupervised model-obtained results, these patterns are not particularly useful for coaches or performance analysts since they merely reflect common, repeated patterns rather than interesting patterns.
The supervised approach with SPP, by using sequences representing passages of play labelled with points scoring outcomes, by virtue of the computed weights, is able to provide a measure of the importance of patterns of plays to these outcomes.
In addition, compared to the unsupervised models, the supervised SPP model obtained a greater variety of patterns of play, i.e., not only those containing breakdowns and or phases, and also discovered more sophisticated patterns.
%
%%%

\section*{Discussion}
\label{sec-discussion}
%%%
In this study, a supervised sequential pattern mining model called safe pattern pruning (SPP) was applied to data from professional rugby union in Japan, consisting of sequences in the form of passages of play that are labelled with points scoring outcomes. 
The obtained results suggest that the SPP model was useful in detecting complex patterns (patterns of play) that are important to scoring outcomes. 
SPP was able to identify relatively sophisticated, discriminative patterns of play, which make sense in terms of their interpretation, and which are potentially useful for coaches and performance analysts for own- and opposition-team analysis in order to identify vulnerabilities and tactical opportunities.
%%%

%%%
By considering both the scoring and conceding perspectives of the team, insight was able to be obtained that would be useful to both the team as well as opposition teams that are due to play the team.
For both the team and their opposition teams during the 2018 season, linebreaks were found to be most associated with scoring.
For both the team and their opposition teams, lineouts were found to be more beneficial to generate scoring opportunities than scrums.
These results are consistent with \cite{coughlan2019they}, who found that lineouts followed by a driving maul are common approaches to scoring tries (albeit in a different competition, Super Rugby), and with \cite{sasaki2007scoring}, who found that around one-third of tries came from lineouts in the Japan Top League in 2003 to 2005---the highest of any try source.
As well as creating lineouts or perhaps prioritising them over scrums, for opposition teams playing the team, effective strategies may include maintaining possession with repeated phase-breakdown play (by aiming for over six repetitions), shutting down the team's ability to regain kicks, and making sure to find touch on exit plays from kick restarts made by the team.
%%%

%%%
As mentioned, compared to the unsupervised models, the supervised SPP model obtained a greater variety of patterns that were also more complex. 
This is likely due to the advantage of the supervised (i.e., labelled) nature of SPP as well as the safe screening and pattern pruning mechanisms of SPP, which prune out irrelevant sequential patterns and model weights in advance.
%%%

%%%
The approach highlighted the potential utility of supervised sequential pattern mining as an analytical framework for performance analysis in sport, and more specifically, the potential usefulness of sequential pattern mining techniques for performance analysis in rugby.
Although the results obtained are encouraging, a limited amount of data from one sport was used.
Also, spatial information such as field position was not available in the data, which may have improved the analysis.
Although the team that performed a particular event was used in our analysis, which player performed particular events was not considered. This may be interesting to investigate in future work.
A limitation of SPP is that, although we considered the order of events within the sequences and their label, the method does not consider the order of sequences within matches, which could also be of informative value (e.g., a particular pattern occurring in the second half of a match may be more important than if it occurs in the first half).
Furthermore, although SPP was useful for the specific dataset in this study, its usefulness is to some degree dependent on the structure of the input data and the specific definition of the sequences and labels. 
For instance, applying the approach to a dataset that consists of entire matches as sequences and win/loss outcomes as the labels does not tend to produce interesting results since it is self-evident that sequences that contain more scoring events will be more associated with wins, thus, SPP would pick up the scoring events on such datasets.
In future work, it would be interesting to apply the approach to a larger amount of data from rugby, as well as to similarly structured datasets in other sports in order to confirm its efficacy.
%%%

\paragraph*{S1 Dataset.}
\label{S1_Dataset}
The delimited sequence data that is described in the paper is available on GitHub: \url{https://github.com/rbun013/Rugby-Sequence-Data}.

\paragraph*{S2 Appendix.}
\label{S2_Appendix}
{\bf Safe Screening and Regularization Path Initialization.}
Some weights are removed prior to solving (\ref{eq:L2-SVM}) using safe screening, which corresponds to finding $j$ such that $w_j = 0$ in the optimal solution $\bw^* := [w_1^*, \ldots, w_{d}^*]^\top$ in the optimization problem (\ref{eq:L2-SVM}).
Such $w_j$ do not affect the optimal solution even if they are removed prior.
In the optimal solution, the $\bw^*$ of the optimization problem {\rm (\ref{eq:LWSM-optimization})}, a set of $j$ such that $|w_j^*| > 0$ is called the active set, and is denoted as $\cA \subseteq \cQ$.
In this case, even if only the subsequence patterns included in $\cA$ are used, the same optimal solution as when using all the subsequence patterns can be obtained.
Thus, if one solves
	\begin{align}
		(\bw^{\prm *}_{\cA},b^{\prm *}) := \argmin_{\bw,b}  &
		\sum_{i \in [n]} \ell(y_i, f(\bg_i ; \{ \bq \}_{i \in \cA})) + \lambda \| \bw \|_1,
		\label{eq:primal-problem-active}
	\end{align}
then it is guaranteed that $\bw^* = \bw^{\prm *}_{\cA}$ and $b^* = b^{\prm *}$.
\\
In practice, 
the $\lambda$ parameter is found
based on a model selection technique 
such as cross-validation. 
In model selection, 
a sequence of solutions, a so-called regularization path, with various penalty parameters must be trained. 
The regularization path of the problem
\eq{eq:LWSM-optimization}, \{$\lambda_{0},\lambda_{1}, \ldots, \lambda_K$\},
is usually computed with decreasing $\lambda$
because sparser solutions are obtained for larger $\lambda$. 

The initial values for computing the regularization path are set to $\bm w^* \xleftarrow[]{} \bm 0$,
$b^* \xleftarrow[]{} \bar{y}$ (where $\bar{y}$ is the sample mean of $\{y_i\}_{i \in [n]}$) and $\lambda_{\rm 0} \xleftarrow[]{} \lambda_{\rm max}$ (see \cite{sakuma2019efficient} for how $\lambda_{\rm max}$ is calculated and for further details of the safe pattern pruning model and its safe-screening mechanism).
%%%

\section*{Acknowledgments}
This work was partially supported by MEXT KAKENHI (20H00601, 16H06538), JST CREST (JPMJCR1502), and RIKEN Center for Advanced Intelligence Project.

\nolinenumbers

% Either type in your references using
% \begin{thebibliography}{}
% Text
% \end{thebibliography}
%
% or
%
% Compile your BiBTeX database using our plos2015.bst
% style file and paste the contents of your .bbl file
% here. See http://journals.plos.org/plosone/s/latex for 

\end{document}